\documentclass[sigconf,anonymous=false,review=false]{acmart}

\usepackage{algorithm}
\usepackage{algorithmic}
\usepackage{multirow}
\usepackage{enumitem}
\usepackage{caption}
\usepackage{booktabs}
\usepackage{amsmath,amsfonts}
\usepackage{textcomp}
\usepackage{subcaption}
\usepackage{footnote}
\usepackage{graphicx}
\usepackage{multirow}
\usepackage{graphicx}
\usepackage{enumitem}
  
\allowdisplaybreaks[4]
\usepackage{hyperref}
\theoremstyle{definition}

\newcommand\M{EGDB}

\newcommand{\tcr}{\textcolor{red}}
\renewcommand\footnotetextcopyrightpermission[1]{}

\AtBeginDocument{%
  \providecommand\BibTeX{{%
    \normalfont B\kern-0.5em{\scshape i\kern-0.25em b}\kern-0.8em\TeX}}}

\copyrightyear{2025} 
\acmYear{2025} 
\setcopyright{acmlicensed}\acmConference[CIKM '25]{Proceedings of the 34th
ACM International Conference on Information and Knowledge
Management}{November 10--14, 2025}{Seoul, Republic of Korea}
\acmBooktitle{Proceedings of the 34th ACM International Conference on
Information and Knowledge Management (CIKM ‘25), November 10--14, 2025,
Seoul, Republic of Korea}
\acmDOI{10.1145/3746252.3761497}
\acmISBN{979-8-4007-2040-6/2025/11}

\begin{document}

\title{Expert-Guided Diffusion Planner for Auto-bidding}
\author{Yunshan	Peng}
\authornote{All authors contributed equally to this research.}
\affiliation{
  \institution{Kuaishou Technology, Beijing, China}
  \city{}
  \country{}
}
\email{pengyunshan@kuaishou.com}

\author{Wenzheng Shu}
\authornotemark[1]
\affiliation{
  \institution{Kuaishou Technology, Beijing, China}
  \city{}
  \country{}
}
\email{shuwenzheng@kuaishou.com}

\author{Jiahao Sun}
\authornotemark[1]
\affiliation{
  \institution{Xi'an Jiaotong University}
  \city{}
  \country{}
}
\email{sunjiahao@stu.xjtu.edu.cn}

\author{Yanxiang Zeng}
\authornotemark[1]
\affiliation{
  \institution{Kuaishou Technology, Beijing, China}
  \city{}
  \country{}
}
\email{zengyanxiang@kuaishou.com}

\author{Jinan Pang}
\affiliation{
  \institution{Kuaishou Technology, Beijing, China}
  \city{}
  \country{}
}
\email{pangjinan@kuaishou.com}

\author{Wentao Bai}
\affiliation{
  \institution{Kuaishou Technology, Beijing, China}
  \city{}
  \country{}
}
\email{baiwentao@kuaishou.com}

\author{Yunke Bai}
\affiliation{
  \institution{Kuaishou Technology, Beijing, China}
  \city{}
  \country{}
}
\email{baiyunke@kuaishou.com}

\author{Xialong Liu}
\affiliation{
  \institution{Kuaishou Technology, Beijing, China}
  \city{}
  \country{}
}
\email{zhaolei16@kuaishou.com}

\author{Peng Jiang}
\authornote{Corresponding Author.}
\affiliation{
  \institution{Kuaishou Technology, Beijing, China}
  \city{}
  \country{}
}
\email{jiangpeng@kuaishou.com}

\renewcommand{\shortauthors}{Yunshan Peng et al.}

\begin{abstract}
Auto-bidding is widely used in advertising systems, serving a diverse range of advertisers. Generative bidding is increasingly gaining traction due to its strong planning capabilities and generalizability. Unlike traditional reinforcement learning-based bidding, generative bidding does not depend on the Markov Decision Process (MDP), thereby exhibiting superior planning performance in long-horizon scenarios. Conditional diffusion modeling approaches have shown significant promise in the field of auto-bidding. However, relying solely on return as the optimality criterion is insufficient to guarantee the generation of truly optimal decision sequences, as it lacks personalized structural information. Moreover, the auto-regressive generation mechanism of diffusion models inherently introduces timeliness risks. To address these challenges, we introduce a novel conditional diffusion modeling approach that integrates expert trajectory guidance with a skip-step sampling strategy to improve generation efficiency. The efficacy of this method has been demonstrated through comprehensive offline experiments and further substantiated by statistically significant outcomes in online A/B testing, yielding an 11.29\% increase in conversions and a 12.36\% growth in revenue relative to the baseline.
\end{abstract}

\ccsdesc[500]{Information systems~Recommender systems}

\keywords{Auto Bidding, Generative Learning, Diffusion Model}

\maketitle
\section{Introduction}
The bidding mechanism of online advertising platforms, as a core component of modern digital marketing \cite{yuan2013real,wang2017display,adikari2015real,evans2009online}, remains pivotal in enabling businesses to target audiences and drive commercial outcomes precisely. However, the inherent volatility of dynamic market conditions and the intricate diversity of user behaviors, combined with the need to process trillions of daily ad impressions in real time \cite{guo2024aigb, johnson2020consumer}, expose dual challenges to traditional manual bidding strategies. Advertisers must continually adjust their bid prices to stay competitive in a rapidly evolving bidding environment, which demands extensive knowledge in the field. Still, they frequently do not reach the best results. 

To address these challenges, advertising platforms have developed auto-bidding services \cite{wen2022cooperative, aggarwal2024auto, lin2024robust} to simplify bid optimization for advertisers. Through carefully designed bidding strategies that account for various advertising factors (such as impression distributions, budgets, and cost constraints), these services automatically set bid prices for each impression opportunity. Various advanced technologies have recently been adopted to meet these demands, including Reinforcement Learning (RL)-based and generative methods. Reinforcement learning models the bidding process as a Markov Decision Process (MDP) and has been widely adopted in automated bidding systems, demonstrating consistent performance enhancements across diverse advertising scenarios \cite{he2021unified,lin2023safe,mou2022sustainable}. However, conventional reinforcement learning approaches grounded in Markov conditional independence assumptions and dynamic programming principles exhibit inherent limitations in bidding effectiveness, mainly stemming from (i) severe cumulative error propagation due to extended decision horizons in bidding sequences and (ii) ignoring the impact of long-term dependencies in the bidding environment \cite{zhu2023diffusion}. Generative bidding demonstrates superior long-term planning capabilities and has gained significant traction.\cite{li2024gas} applies the Decision Transformer to conduct sequential planning. \cite{guo2024generative} applies a diffusion planner, incorporating influences from preceding states and enabling extended-horizon strategic planning. 

However, applying conditional diffusion models to bidding faces two key challenges. First, existing approaches condition cumulative returns as a single scalar hyperparameter during inference, leading to weak guidance due to (i) \textit{lack of personalization} and (ii) \textit{vast solution space satisfying the scalar constraint}, thereby hindering convergence to optimal solutions. Second, the online bidding module’s strict latency requirements conflict with the T-step iterative sampling process inherent to diffusion models. To tackle these challenges discussed previously, we introduce a novel framework called \textbf{E}xpert-\textbf{G}uided \textbf{D}iffusion planner for Auto-\textbf{B}idding (\M). Firstly, we propose a \textbf{blended-forcing mechanism} that synergistically integrates teacher-forcing stabilization and VAE-enhanced decode forcing. It enables precise imitation of expert trajectories while proactively exploring latent optimal bidding patterns beyond observed demonstrations. Secondly, building upon the refined expert guidance, we devise a \textbf{dual-conditioned planner} that implicitly incorporates expert behavioral semantics while explicitly enforcing domain constraints (i.e., budgets), achieving principled adaptation to dynamic auction environments. Finally, to overcome the inherent latency of iterative diffusion, we develop a \textbf{step-skipping strategy} that reduces inference steps without compromising the quality of action generation — a crucial advancement for real-time bidding systems. We demonstrate how this framework achieves state-of-the-art offline and online performance through comprehensive experiments. 

In conclusion, our main contributions can be summarized in the following four parts:
\begin{itemize}[leftmargin=*]
    \item We construct expert demonstrations explicitly and leverage them to provide personalized guidance for the diffusion generation process.
    \item To ensure training-inference consistency, we introduce a Variational Auto-Encoder to model expert inherent behaviors, coupled with blended-forcing techniques to mitigate error accumulation and enhance training efficiency. 
    \item We adopt accelerated sampling with skip-step to further speed up the generation process.
    \item To verify the effectiveness of the proposed method, we conduct extensive offline experiments and perform ablation studies to assess the contribution of each component. Moreover, we demonstrate its practical efficacy through online deployment.
\end{itemize}

\section{Related Work}
\subsection{Offline Reinforcement Learning}
In recent years, Offline reinforcement learning (RL) has emerged as a pivotal paradigm for decision-making. It aims to learn optimal policies solely from static datasets of historical interactions, without requiring online environment exploration. This paradigm addresses scenarios where allowing partially trained agents to interact with the environment could lead to safety risks or substantial costs. Offline RL methods \cite{iql, bcq, cql, dt} address two critical challenges—distribution shift and extrapolation errors-through distinct methodological innovations. Implicit Q-Learning (IQL) \cite{iql} tackles distribution shift by implicitly estimating optimal actions through upper expectile regression on state values, suppressing extrapolation errors via asymmetric Q-updates that prioritize in-distribution high-value actions. Batch-Constrained Q-learning (BCQ) \cite{bcq} combines a variational autoencoder (VAE) with Q-filtering to restrict policy updates to data-supported regions, minimizing overestimation risks. Conservative Q-Learning (CQL) \cite{cql} explicitly regularizes Q-values with theoretical guarantees, penalizing out-of-distribution (OOD) actions through conservative value bounds. In contrast, Decision Transformer (DT) \cite{dt} circumvents value functions entirely by reframing RL as sequence modeling, autoregressively generating actions conditioned on historical trajectories to eliminate bootstrapping errors and OOD queries.

\subsection{Diffusion Model for Decision Making}
Diffusion models \cite{ho2020denoising, song2020denoising}, built upon stochastic denoising processes, have emerged as a dominant paradigm in image synthesis, capable of generating high-fidelity data through Markov chain-based iterative refinement. These models can be divided into two categories: conditional and unconditional paradigms. Conditional diffusion models \cite{fu2024unveil} incorporate additional information (e.g., class labels, text prompts, or environmental states) to steer the generation process, enabling targeted synthesis. While unconditional diffusion models \cite{heng2024out} generate data from pure noise without external guidance, excelling in capturing the intrinsic distribution of the training data. Owing to the powerful capabilities of diffusion models, they have been widely applied across various fields, particularly in decision-making. \cite{hansen2023idql} proposes IDQL, an RL framework that integrates implicit Q-learning with diffusion policies in an actor-critic architecture. \cite{wang2022diffusion} leverages diffusion models to capture complex, multimodal behavior distributions from static datasets. \cite{chen2022offline} uses diffusion models to more accurately fit the complex distribution of behavioral policies and avoids selecting unseen actions that are out of the data distribution during training.

\subsection{Real-Time Auto-Bidding}
Auto-Bidding in online advertising dynamically adjusts bidding strategies to maximize effectiveness (e.g., clicks, conversions, or sales) under constraints such as budgets and return on ad spend (ROAS). RL-based methods are widely used in auto-bidding because they dynamically optimize bids in complex advertising environments and balance exploration and exploitation. \cite{cai2017real} proposes an RL-based framework to optimize real-time bidding strategies in display advertising. \cite{he2021unified} proposes a generalized framework for optimizing real-time bidding strategies under diverse budget and key performance indicator (KPI) constraints, dynamically adjusting bid parameters via RL to maximize campaign value while ensuring compliance with advertiser requirements. \cite{jin2018real} explores multi-agent RL for real-time bidding in display advertising. Recent studies \cite{gao2025generative, guo2024generative} have increasingly explored generative auto-bidding strategies in online advertising. \cite{gao2025generative} leverages Decision Transformer \cite{dt} to model sequential dependencies and historical auction dynamics, enabling adaptive bidding strategy optimization by autoregressively generating context-aware bid adjustments aligned with long-term advertiser objectives. \cite{guo2024generative} directly models the correlation between the return and the entire trajectory and uses a conditional diffusion for bid generation. 

\section{Preliminary}
In this section, we outline the general bidding and auction tasks framework and examine the architecture of generative diffusion models. To prevent any confusion regarding subscripts in the subsequent discussion, it should be noted that $k$ denotes the diffusion time-steps, while $t$ represents the position within the trajectory $\tau$.

\subsection{Formulation of Auto-Bidding Problem}
\label{subsec:bidding}
In online advertising auctions, the bidding problem typically involves maximizing a specified objective while adhering to certain constraints. For each advertiser $i$, the most frequently examined are budget ($B_i$) and \textit{return-on-spend} (\textit{RoS}) constraints ($C_i$). Consequently, the bidding problem can be formulated as follows~\cite{he2021unified}:
\begin{align}
\begin{aligned}
    \label{eq:1}
    \max \sum_{j} \mathbb{A}_{ij} \cdot v_{ij} \quad s.t. \sum_{j} (\mathbb{A}_{ij} \cdot b_j) &\leq B_i,\\
    \frac{\sum_{j} (\mathbb{A}_{ij} \cdot \mathbb{P}_{ij})}{\sum_{j} (\mathbb{A}_{ij} \cdot v_{ij})} &\leq C_{i}.
\end{aligned}
\end{align}
Here, $b_j$ denotes the bids submitted by all advertisers for each auction $j$, and $v_{ij}\in R^+$ signifies the revenue generated by the $i$-th advertiser through auction $j$. The term $\mathbb{A}_{ij}(b_{ij})\in[0,1]$ indicates whether advertiser $i$ wins auction $j$, and $\mathbb{P}_{ij}$ represents the payment made by advertiser $i$ to secure auction $j$.

\begin{figure*}[t]
  \centering
  \includegraphics[width=\textwidth]{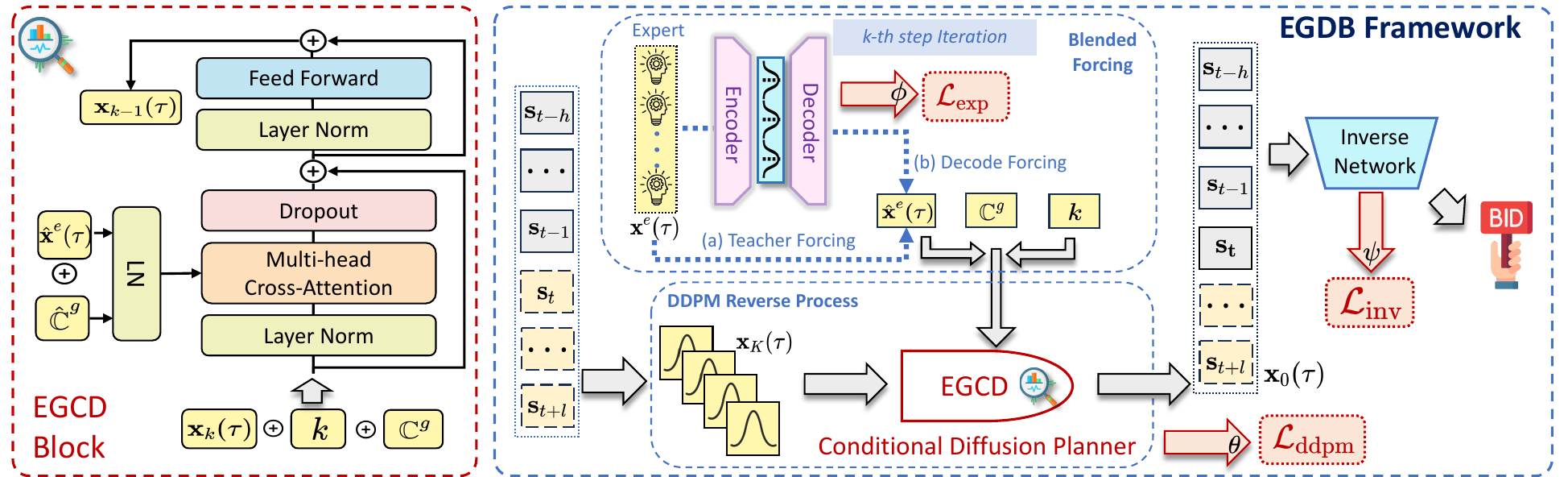}
  \caption{The foundational framework of our proposal consists of the following components: (1) the block of expert-guided conditional diffusion (EGCD) and (2) the overall architecture of ~\M.}
  \label{fig:framework}
\end{figure*}

\subsection{Basic Diffusion Model}
\label{subsec:dm}
The diffusion model (DM)~\cite{ho2020denoising} functions via forward and reverse processes, and it is widely used in recommender systems. Starting with an initial data sample $x_0\sim q(x_0)$, the forward process incrementally introduces Gaussian noise through a Markov chain over $K$ steps, resulting in latent variables $x_{1:K}$. The forward transition is defined by $q(x_k|\boldsymbol{x}_{k-1})=\mathcal{N}(\boldsymbol{x}_k;\sqrt{1-\beta_k}\boldsymbol{x}_{k-1},\beta_k\boldsymbol{I})$, where $k\in\{1,\cdots,K\}$ denotes the diffusion step, $\mathcal{N}$ represents the Gaussian distribution, and $\beta_k\in(0,1)$ controls the noise intensity at step $k$. As $K$ approaches infinity, $x_K$ converges to a standard Gaussian distribution. The DM reconstructs $x_{k-1}$ from $x_k$ by learning a denoising trajectory in the reverse process. Commencing from $x_K$, it progressively approximates $x_k\to\boldsymbol{x}_{k-1}$ through $p_\theta(\boldsymbol{x}_{k-1}|\boldsymbol{x}_k)=\mathcal{N}(\boldsymbol{x}_{k-1};\mu_\theta(\boldsymbol{x}_k,k),\Sigma_\theta(\boldsymbol{x}_k,k))$, where $\mu_\theta(x_k,k)$ and $\Sigma_\theta(x_k,k)$ are outputs of a neural network parameterized by $\theta$, capturing the intricate generative dynamics.

\subsection{Diffusion with Classifier-Free Guidance}
\label{subsec:df}
The sequential nature of auto-bidding requires capturing both temporal dependencies and constraint satisfaction (e.g., budget limits). Traditional autoregressive methods often suffer from error accumulation in multi-step generation, while standard generative models struggle to incorporate dynamic constraints. Consequently, following \cite{guo2024aigb}, we formulate the sequential decision problem through conditional generative modeling:
\begin{equation}
    \max_\theta \mathbb{E}_{\tau \sim D} \left[ \log p_\theta \left( x_0(\tau) \mid y(\tau) \right) \right],
\end{equation}
where $\tau$ denotes a trajectory identifier, $x_0(\tau)$ is the original trajectory of states$(s_1, ..., s_t, ..., s_T)$ and $y(\tau)$ encapsulates auxiliary conditions. The goal is to estimate the conditional data distribution with $p_\theta$ so that the future states of a trajectory $x_0(\tau)$ from information $y(\tau)$ can be generated. For example, in online advertising, $y(\tau)$ can be the constraints or the total value of the entire trajectory. Under such a setting, we can formalize the conditional diffusion modeling for auto-bidding:
\begin{equation}
    q\left( x_{k+1}(\tau) \mid x_k(\tau) \right), \quad p_\theta \left( x_{k-1}(\tau) \mid x_k(\tau), y(\tau) \right),
\end{equation}
where $q$ represents the forward process in which noises are gradually added to the trajectory, while $p_\theta$ is the reverse process, where a model is used for denoising.

We model the forward process $q\left( x_{k+1}(\tau) \mid x_k(\tau) \right)$ via diffusion over states:
\begin{equation}
    x_k(\tau) := \left( s_1, \dots, s_t, \dots, s_T \right)_k,
\end{equation}
where $s_t$ is a one-dimensional vector, including real-time bidding context, and $x_k(\tau)$ is a noise sequence of states which can be represented by a two-dimensional array where the first dimension is the periods. The second dimension is the state values. Given $x_k(\tau)$, the forward process of the diffusion model can be represented as follows:
\begin{equation}
    q\left( x_k(\tau) \mid x_{k-1}(\tau) \right) = \mathcal{N} \left( x_k(\tau); \sqrt{1 - \beta_k} x_{k-1}(\tau), \beta_k \mathbf{I} \right).
\end{equation}
When $k \to \infty$, $x_k(\tau)$ approaches a sequence of standard Gaussian distribution, where we can make sampling through the reparameterization trick and then gradually denoise the trajectory to produce the final state sequence.

The classifier-guided diffusion model requires an accurate estimation of the guidance gradient based on the trajectory classifier $y(\tau)$, which may not be feasible. Therefore, we use a classifier-free guidance strategy to guide the generation. The core of Classifier-Free Guidance lies in replacing explicit classifiers with an implicit classifier, eliminating the need to compute explicit classifiers and their gradients directly. Following \cite{ajay2022conditional,ho2022classifier}, who pioneered conditional guidance in diffusion models, we utilize a classifier-free guidance strategy with low-temperature sampling. During the training phase, we need to train the unconditional model $\epsilon_\theta \left( x_k(\tau), k \right)$ and the conditional model $\epsilon_\theta \left( x_k(\tau), y(\tau), k \right)$. However, these can be represented by a single shared model by randomly dropping the condition. The final output is obtained through a linear extrapolation of the conditional and unconditional generation predictions:
\begin{equation}
    \hat{\epsilon}_k := \epsilon_\theta \left( x_k(\tau), k \right) + \omega \left( \epsilon_\theta \left( x_k(\tau), y(\tau), k \right) - \epsilon_\theta \left( x_k(\tau), k \right) \right),
\end{equation}
where the guidance coefficient $\omega$ can be adjusted to control the trade-off between the realism and diversity of generated samples. After that, we can sample from $p_\theta \left( x_{k-1}(\tau) \mid x_k(\tau), y(\tau) \right)$ as follows:
\begin{equation}
    x_{k-1}(\tau) \sim \mathcal{N} \left( x_{k-1}(\tau) \mid \mu_\theta \left( x_k(\tau), y(\tau), k \right), \Sigma_\theta \left( x_k(\tau), k \right) \right),
\end{equation}
when serving at period $t$, we first sample from an initial trajectory $x’_k(\tau) \sim \mathcal{N}(0, \mathbf{I})$ and assigns the history states $s_{0:t}$ into it. Then, we can sample predicted states with the reverse process recursively by
\begin{equation}
    x’_{k-1}(\tau) = \mu_\theta \left( x’_k(\tau), y(\tau), k \right) + \sqrt{\beta_k} z,
    \label{eq:10}
\end{equation}
where $z \sim \mathcal{N}(0, \mathbf{I})$. Given $x'_0(\tau)$, we can extract the next predicted state $s'_{t+1}$, and determine how much it should bid to achieve that state. In this setting, we apply the widely used inverse dynamics \cite{agrawal2016learning}  with a non-Markov state sequence to determine current bidding parameters at period $t$:
\begin{equation}
    \hat{a}_t = f_\phi \left( s_{t-h:t}, s’_{t+1} \right),
    \label{eq:11}
\end{equation}
where $\hat{a}_t$ contains predicted bidding parameters at time $t$ and $h$ is the length of history states.

\section{Methodology}
In this section, we present a comprehensive description of our proposed \M~ framework. We begin by formalizing the core mechanism of Blended-Forcing with optimal experts in Section \ref{subsec:blend}. Next, in Section \ref{subsec:egcd}, we elaborate on the expert-guided conditional planner that translates blended signals into structured action sequences. Finally, Section \ref{subsec:generation} details the efficient action generation module designed for real-time execution. The overall architecture is illustrated in Figure \ref{fig:framework}.

\subsection{Blended-Forcing with Optimal Expert}
\label{subsec:blend}
To address the bidding problem outlined in Equation \ref{eq:1}, we initially aim to generate expert trajectories. To concentrate on a single bidding agent, we remove the subscript $i$, assuming all other agents are fixed. We transform the linear programming (LP) function into its dual form. Assuming the auction is truthful, the bidding formula will produce an outcome corresponding to the
optimal primal solution $x_j^e$~\cite{aggarwal2024auto}. For each query $j$:
\begin{align}
\label{eq:2}
\mathbf{x}^e(\tau)= (x_1^e,\ldots,x_T^e), \quad x_j^e = \frac{1 + \alpha_{c} \cdot C_i}{\alpha_{b} + \alpha_{c}} \cdot v_j,
\end{align}
where $\alpha_{b}$ represents the influence of budget constraints and $\alpha_{c}$ denotes the impact of constraints of cost. Then, we can refine the $\mathbf{x}^e(\tau)$ to fully capture the latent expert behaviors through the blended-forcing mechanism in the training phase. As shown in Figure \ref{fig:framework}, it comprises two operations: (i) the teacher-forcing technique ($TF(\cdot)$) to imitate the expert behaviors more stably, and (ii) the decode-forcing mechanism ($DF(\cdot)$) to explore more optimal expert behaviors through an extra Variational Auto-Encoder (VAE)\cite{kingma2013auto}.

Teacher-Forcing (TF), initially introduced in Recurrent Neural Networks (RNNs) \cite{williams1989learning}, is a training technique where the ground-truth sequence is fed as input to the model at each time step, rather than using its previous predictions. This approach effectively mitigates error accumulation in autoregressive models by aligning training dynamics with the actual data distribution. However, a critical limitation arises during inference: the absence of ground truth forces the model to rely entirely on its own (potentially erroneous) predictions, resulting in a mismatch between the training and inference input distributions. This phenomenon, termed Exposure Bias, often results in degraded performance in real-world scenarios. Inspired by this technique, we introduce TF and use expert trajectories serving as ground truth sequences to help the model imitate expert behaviors. 

While TF alleviates error propagation during training, its reliance on ground truth introduces a fundamental inconsistency between the training and inference phases. In complex dynamic environments (e.g., real-time bidding), traditional mitigation strategies like Scheduled Sampling \cite{bengio2015scheduled} — which heuristically blends ground truth and predicted tokens — fail to capture rapidly shifting environmental dynamics adaptively. The rigidity of such handcrafted mixing strategies often leads to suboptimal policy generalization, particularly when expert-guided constraints (e.g., budget allocation rules) must be preserved.

To bridge this gap, we propose a VAE-enhanced hybrid mechanism called \textbf{Blended-Forcing}, which implicitly models environmental dynamics through the lens of expert historical behaviors. The VAE’s latent space is trained to encode the distribution of optimal expert actions conditioned on environmental states. During inference, the VAE reconstructs a pseudo ground-truth action that approximates expert-level decisions given the current state, thereby serving as a dynamically adaptive substitute for the missing ground truth.

Moreover, we apply for coverage $\mathbf{x}^e(\tau)$ to $\hat{\mathbf{x}}^e(\tau)$ with a hyperparameter $\delta$ to harmonize the representation of the expert, which can be formulated as:
\begin{align}
\label{eq:3}
\hat{\mathbf{x}}^e(\tau) = \mathbb{I}[TF(\mathbf{x}^e(\tau))_{\delta},DF(\mathbf{x}^e(\tau))_{1-\delta}],
\end{align}
where $\mathbb{I}[\cdot]$ represents the indicator function, and in each $k$-th diffusion step, it will choose $TF(\cdot)$ in the probability of $\delta$ and $DF(\cdot)$ otherwise. As for the structure of $DF(\cdot)$, we implement a one-layer DNN for both the encoder and decoder with the learning parameters $\phi$ for reconstruction.

\subsection{Expert-Guided Conditional Planner}
\label{subsec:egcd}
After we obtain the $\hat{\mathbf{x}}^e(\tau)$ as the extra condition for diffusion training, it serves as the implicit signal during the inference phase. Consequently, the overall framework should be guided more explicitly. Therefore, we incorporate the expected constraints $\mathbb{C}^g$ defined in Section \ref{subsec:bidding} to formulate the overall condition $\mathbb{C}$ in a dual fashion, which can be depicted as:
\begin{align}
\label{eq:4}
\mathbb{C}=\underbrace{\hat{\mathbf{x}}^e(\tau)}_{IMPLICIT} \oplus \quad \underbrace{[f_{\text{}}(R(\tau) \oplus f_{\text{}}’ (C(\tau))]}_{EXPLICIT (\mathbb{C}^g)}.
\end{align}
Note that $R(\tau)$ denotes the total value the advertiser received as the return $R(\tau)=\sum_{t=1}^{T}r_{t}$. $f$ is min-max normalization,  $f'$ denotes a indicator function   transfers constraint to $[0,1]$ corresponding to \cite{guo2024generative}, and $\oplus$ represents the concatenation operation. Since DMs are optimized by maximizing the Evidence Lower Bound (ELBO) of the likelihood of observed input data~\cite{ho2020denoising}, we adapt the optimization objective $\mathcal{L}_{\text{ddpm}}$ stated in Section \ref{subsec:dm} previously as:
\begin{equation}
    \label{eq:5}
    D_{\text{KL}}(q(\mathbf{x}_{k-1}(\tau)\mid\mathbf{x}_k(\tau),\mathbf{x}_0(\tau))\mid\mid p_\theta(\mathbf{x}_{k-1}(\tau)\mid\mathbf{x}_k(\tau),\mathbb{C})).
\end{equation}
In line with prior researches ~\cite{li2023diffurec,yang2024generate}, we utilize an alternative
parameterization trick that predicts the target $\mathbf{x}_0$ rather than the added Gaussian noise $\epsilon$, to circumvent the representation collapse issue ~\cite{wang2024conditional} in discrete space as:
\begin{align}
    \label{eq:6}
    \tilde{\boldsymbol{\mu}}_\theta(\mathbf{x}_k(\tau),\mathbb{C},k)=\sqrt{\bar{\alpha}_{k-1}}f_\theta(\mathbf{x}_k(\tau),\mathbb{C},k)+\frac{(1-\bar{\alpha}_{k-1})\sqrt{\alpha_k}}{\sqrt{1-\bar{\alpha}_k}}\boldsymbol{\epsilon}.
\end{align}
Drawing inspiration from the success of sequential recommendation models in ~\cite{huang2024dual}, we designed a novel block named Expert-Guided Conditional Diffusion (EGCD) for the diffusion model architecture. This model is specifically built on the Transformer architecture~\cite{vaswani2017attention}, where we have substituted the traditional self-attention mechanism with cross-attention to leverage the conditional information fully.

\begin{algorithm}[t]
\caption{Training with EGDB}
\label{algo:training}
\begin{algorithmic}[1] 
\STATE \textbf{Inputs:} diffusion step $K$, expert trajectory $\boldsymbol{x}^{e}(\tau)$.
\STATE \textbf{Parameters:} $\Theta=\{\phi, \theta, \psi\}$
\STATE \textbf{repeat}
\STATE \quad Attain $\hat{\mathbf{x}}^e(\tau)$ through Eq. (\ref{eq:3}) with probability $\delta$
\STATE \quad Calculate $\mathbb{C}$ through Eq. (\ref{eq:4})
\STATE \quad Sample $k \sim \text{Uniform}\{1,\cdots,K\}$ and $\boldsymbol{x}_k(\tau)[:t]\leftarrow \mathbf{s}_{0:t}$
\STATE \quad Estimate $\hat{\textbf{x}}_0(\tau) \leftarrow (\tilde{\mu}_{k-1},\Sigma_{k-1})$ through Eq. (\ref{eq:5})
\STATE \quad $\boldsymbol{x}_{k-1}(\tau)\sim\mathcal{N}(\tilde{\mu}_{k-1},\alpha\Sigma_{k-1})$
\STATE \quad Take gradient descent step $\nabla_\Theta(\mathcal{L}_{\text{total}})$ through Eq. (\ref{eq:8})
\STATE \textbf{until} converged
\end{algorithmic}
\end{algorithm}

To formulate the query for cross-attention, we concatenate the original sequence generated in a single step, denoted as  $x_k(\tau)$, the total number of denoising steps $k$, and the relevant conditions $\mathbb{C}^g$ as the query:

\begin{equation}
    \label{eq:12}
    \mathbb{O} = x_k(\tau)\oplus k \oplus \mathbb{C}^g,\ \text{and} \ \mathbb{Q} = \texttt{L-Norm}[\text{MLP}(\mathbb{O})].
\end{equation}
It is noted that $\texttt{L-Norm}[\cdot]$ and $\text{MLP}(\cdot)$ represent the layer normalization and dense layer, respectively, and cross-attention is defined as the process of extracting effective information from the expert sequence as follows:

\begin{align}
    \label{eq:13}
    \text{Cross-Attn}=\texttt{Softmax} \left( \frac{(\mathbb{Q} W^Q)(\mathbb{C}W^K)^T}{\sqrt{d}}\right) (\mathbb{C} W^V),
\end{align}
where the Key and Value of cross-attention is defined in \eqref{eq:4} and the parameter matrices $W^Q$,$W^K$,$W^V$ represent the weights associated with the transformations of the original $Q$,$ K$, and $V$, and $d$ denotes the dimension after the vectors have been concatenated. The latent representation $\mathbb{H}_0$ can be obtained from cross-attention, and $x_{k-1}(\tau)$ is obtained as follows:

\begin{equation}
    \label{eq:113}
    \mathbb{H}_1 = \text{FFN}(\texttt{L-Norm}[\mathbb{H}_0 + \mathbb{O}]), \ \ x_{k-1}(\tau) = \text{MLP}\left( \mathbb{H}_1 + \mathbb{H}_0 \right),
\end{equation}
where $\text{FFN}(\cdot)$ is the Feed-Forward Network, $\mathbb{H}_0$ and $\mathbb{H}_1$ are add as output.

\subsection{Efficient Action Generation}
\label{subsec:generation}
\subsubsection{Accelerate Sampling.}
By employing Equation \ref{eq:6}, the inference Markov transition of ~\M~ can be defined as
$p_{\theta}(\mathbf{x}_{k-1}(\tau)\mid\mathbf{x}_k(\tau))$. Each step is designed to predict $\hat{\mathbf{x}}_0(\tau)$ and provide $\mathbf{x}_{k-1}(\tau)$, which is less noisy than $\mathbf{x}_k(\tau)$ for the next step. The time cost of the inference process increases linearly with the total number of steps. A natural approach to accelerate this process is to skip several intermediate steps if the early approximations are accurate enough and require no further adjustment. Here, we define $p_\theta(\mathbf{x}_{k-\gamma}(\tau)\mid \mathbf{x}_{k}(\tau)) \rightarrow p_\theta(k-\gamma\mid k)$ with the number of skipped steps $\gamma$ as:
\begin{align}
\label{eq:7}
    p_\theta(k-\gamma\mid k)=\mathcal{N}(\mathbf{x}_{k-\gamma}(\tau);\tilde{\mu}_\theta(\hat{\mathbf{x}}_0(\tau),t),\Sigma_\theta(\hat{\mathbf{x}}_0(\tau),t)),
\end{align}
where the $\gamma$ governs the number of skip steps. This acceleration method significantly reduces the time overhead without sacrificing excessive accuracy. We will discuss the details in Section \ref{subsec:performance}. After we get $\hat{\mathbf{x}}_0(\tau)$, we can generate the corresponding action $\hat{\mathbf{a}}_t$ through an inverse network $f_\psi(\cdot)$.

\begin{algorithm}[t]
\caption{Generating Optimal Actions}
\label{algo:inference}
\begin{algorithmic}[1]
\STATE \textbf{Inputs:} diffusion step $K$, expert trajectory $\mathbf{x}^{e}(\tau)$.
\STATE \textbf{Parameters:} $\Theta=\{\phi, \theta, \psi\}$
\STATE Get history of states $\mathbf{s}_{0:t}$; 
\STATE Sample $\mathbf{x}_{K}(\tau) \sim \mathcal{N} (0, \beta_{K}I)$, and $\hat{\mathbf{x}}^e(\tau) \sim \text{Norm}[0,1]$
\STATE Decode $\hat{\mathbf{x}}^e(\tau)$ with $z$ and calculate $\mathbb{C}$ through Eq. (\ref{eq:4})
\STATE \textbf{repeat}
\STATE \textbf{for} $k = \{K, K-1,\cdots, 1\}$ \textbf{do}
\STATE \quad $\mathbf{x}_k(\tau)[:t]\leftarrow \mathbf{s}_{0:t}$
\STATE \quad Estimate $\hat{\textbf{x}}_0(\tau)\leftarrow(\mu_{k-\gamma},\Sigma_{k-\gamma})$ through Eq. (\ref{eq:7})
\STATE \quad $\mathbf{x}_{k-1}(\tau)\sim\mathcal{N}(\tilde{\mu}_{k-1},\alpha\Sigma_{k-1})$
\STATE \textbf{end for}
\STATE Extract $(\mathbf{s}_{t-h:t},\mathbf{s}'_{t+1})$ from $\hat{\mathbf{x}}_0(\tau)$
\STATE Generate $\hat{\mathbf{a}}_{t}=f_{\psi}(\mathbf{s}_{t-h:t},\mathbf{s}'_{t+1})$;
\end{algorithmic}
\end{algorithm}

\subsubsection{Final Optimization.}
As for optimization, considering that VAE is employed for extracting information from expert sequences, the VAE loss~\cite{kingma2013auto} must also be incorporated as $\mathcal{L}_{\text{exp}}$, which primarily consists of the reconstruction loss of $z$ and the constraint that the intermediate latent variable $z$ follows a normal distribution. Moreover, to inverse actions through the series of states generated by the conditional diffusion planner, we utilize the inverse dynamics loss $\mathcal{L}_{\text{inv}}$ to optimize the reconstruction of the action patterns following~\cite{guo2024generative}. Therefore, the optimization can be simplified as:
\begin{align}
\begin{aligned}
    \label{eq:8}
    \mathcal{L}_{\text{total}}&= \mathcal{L}_{\text{ddpm}} + \xi (\mathcal{L}_{\text{exp}} + \mathcal{L}_{\text{inv}}), \ \text{where} \\
    \mathcal{L}_{\text{ddpm}}&=\mathbb{E}_{\boldsymbol{\epsilon},\mathbf{x}_0(\tau)}||\mathbf{x}_0(\tau)-f_\theta(\sqrt{\bar{\alpha}_k}\mathbf{x}_0(\tau)+\sqrt{1-\bar{\alpha}_k}\boldsymbol{\epsilon},\mathbb{C},k)||^2,\\
    \mathcal{L}_{\text{exp}}&=\frac{1}{2}\left(1+\log{((\sigma_{z})}^{2})-(\mu_{z})^{2}-(\sigma_{z})^{2}\right)+\log p_{\boldsymbol{\theta}}(\mathbf{x}^{e}(\tau)|\mathbf{z}),\\
    \mathcal{L}_{\text{inv}}&=\mathbb{E}_{\mathbf{s}_{t-h:t},\mathbf{a}_t,\mathbf{s}'_{t+1}}||\mathbf{a}_t-f_{\phi}(\boldsymbol{s}_{t-h:t},\boldsymbol{s}'_{t+1})||^2.
\end{aligned}
\end{align}
Here, $\xi$ governs the reconstruction of the expert trajectory and action generation. The overall training and inference phases are shown in Algorithm \ref{algo:training} and  Algorithm \ref{algo:inference}, respectively.

\section{Experiments}
\subsection{Experimental Setting}
We conduct extensive experiments both in offline environments and online A/B testing to demonstrate the effectiveness of our proposed method. Four research questions are investigated in the following experiments:
\begin{itemize}[leftmargin=*]
    \item \textbf{RQ1}: How does \M~ perform compared to state-of-the-art auto-bidding baselines?
    \item \textbf{RQ2}: How do the proposed components in \M~ contribute to the final bidding performance?
    \item \textbf{RQ3}: How do parameter settings impact the model’s performance?
    \item \textbf{RQ4}: How does ~\M~ perform in real-world business scenarios compared to baselines, and is ~\M~ effective in improving inference efficiency?
\end{itemize}

\label{subsec:env}
\subsubsection{Dataset \& Environment.} We utilize Alibaba\footnote{https://github.com/alimama-tech/AuctionNet} simulated environment from the NeurIPS 2024 competition~\cite{AIGB_data}. The dataset covers 21 periods (7-27), each with over 500,000 impression opportunities and 48 steps. Each opportunity involves 48 bidding agents. The dataset contains over 500 million records, each with predicted conversion values, bids, auction details, and impression results. For a fair comparison, data from periods 7-13 is for training and 14-20 for evaluation.

\subsubsection{Metrics.} To simplify the experimental evaluation, we kept the bids of 47 of the 48 advertisers unchanged during the test, only adjusted the bid of the remaining advertiser, and simulated the corresponding rewards and impressions. To evaluate these methods, we use the cumulative conversions of all advertisers, adjusted for penalties incurred by exceeding the Cost Per Action  (CPA) ratio, as our metric.
\begin{align}
\begin{aligned}
    \label{eq:9}
    \text{score} = \mathbb{P}(CPA,C)\sum_i x_i E_i V_i, \ \text{and} \\
    \mathbb{P}(CPA,C) = \text{min}\{(\frac{CPA}{C})^\lambda, 1\}.
\end{aligned}
\end{align}
Here, $x_i$ indicates whether the agent secures the impression, $E_i$ signifies if the ad is successfully shown to the user after winning, and $V_i$ denotes whether a conversion occurs after the user is exposed to the advertisement. The advertiser sets the target $CPA$, while $C$ represents the actual $CPA$ after completing the process. The penalty function defines a hyperparameter $\lambda$, usually set to 2.

\subsubsection{Baselines.}
To verify the effectiveness of our proposed method, we list some common reinforcement learning methods as benchmarks for comparison:

\begin{itemize}[leftmargin=*]
    \item \textbf{CQL}~\cite{cql} addresses overestimation of out-of-distribution actions by introducing conservative constraints in the Q-function.
    \item \textbf{IQL}~\cite{iql} focuses on in-sample actions and avoiding the querying of values for unseen actions.
    \item \textbf{BC}~\cite{bc} is trained through supervised learning by directly replicating expert trajectory data.  
    \item \textbf{DT}~\cite{dt} is a Transformer-based framework that maps trajectories to action outputs through sequence modeling. 
    \item \textbf{DiffBid}~\cite{guo2024aigb} utilizes conditional diffusion modeling approach for bid generation.
    \item \textbf{EBaReT}~\cite{li2025ebaret} integrates the Bag Reward Decision Transformer with expert trajectory generation.
\end{itemize}

\subsection{Experimental Results}
\label{subsec:performance}

\subsubsection{Overall Performances (RQ1)}
We compare the overall performance of ~\M~ with various baselines in Table~\ref{t:overall}. Our method consistently outperforms others across most periods and ranks second in the rest, validating the effectiveness of our proposed approach. This advantage arises from the robust generative capabilities of the diffusion model and the guidance provided by expert samples, which enhances the model’s ability to generate superior bidding trajectories. The IQL and CQL methods outperform BC, suggesting that traditional reinforcement learning can develop more effective strategies than simple supervised learning. Moreover, DT-based methods (e.g., EBaReT) demonstrate superior performance compared to conventional reinforcement learning methods at specific periods, indicating that taking sequence dependency into account is beneficial for optimizing bidding strategies.

\begin{table}[t]
  \caption{Performance comparison between baselines and ~\M. The best performance of each column is bold, while the second-best is \underline{underlined}.}
  \label{t:overall}
  \resizebox{\linewidth}{!}{
  \centering
  \begin{tabular}{lccccccc}
    \toprule
    \textbf{Method}&P14& P15 & P16 & P17 & P18 & P19 & P20 \\
    \midrule
    CQL     & 31.09 &29.99  &29.88  & 29.73  & 27.97  & 33.45  & 27.99     \\
    IQL     & 32.75 & 25.43  & 27.05  & 35.09  & 30.38  & 34.07  & 28.39     \\
    BC & 28.53  & 27.37  & 30.55  & 28.64  & 29.13 & 31.70  & 26.43   \\
    DT     & 30.01 & 29.79  & 30.31  & 31.92  & 29.52  &34.42  & 29.70     \\
    DiffBid     & 31.95 & 28.62  &30.65  &35.59  & 27.72  & 34.36  &29.79    \\
    EBaReT& \underline{37.89} & \underline{33.31}  &\underline{36.09}  &\textbf{40.40}  &\underline{33.42}  &\textbf{40.66}  &\underline{33.71}\\
    \midrule
    \M     &\textbf{38.91} &\textbf{33.69} &\textbf{37.56} &\underline{39.51} &\textbf{34.73} &\underline{38.66} &\textbf{34.01}\\
    \bottomrule
  \end{tabular}
  }
\end{table}

\subsubsection{Ablation Studies (RQ2)}
To further clarify the contributions of each module in ~\M~, we conduct ablation experiments by evaluating various configurations of ~\M. Table~\ref{t:ablation} presents the overall results of our ablation studies on the proposed components, revealing three key insights: (1) All modules contribute positively, with the CA. Block demonstrating a significant impact; (2) An adaptive fast-sampling strategy can achieve optimal performance-accuracy trade-offs;  (3) Blended-Forcing partially emulates expert sequence behaviors, indicating promising regularization potential.

\begin{table}[t]
  \caption{The ablation study results for our proposed ~\M: (1) w/o Acc. $\rightarrow$ the model denoises step by step to replace acceleration; (2) w/o BF. $\rightarrow$ Blended-Forcing is removed and only the Auto-Encoder; (3) w/o CA. $\rightarrow$ Cross-Attention block is removed, and Self-Attention is adopted.}
  \label{t:ablation}
  \resizebox{\linewidth}{!}{
  \centering
  \begin{tabular}{lccccccc}
    \toprule
    \textbf{Method}     & P14& P15 & P16 & P17 & P18 & P19 & P20 \\
    \midrule
    w/o Acc.     & 37.52 & 31.83  & 36.40  & 37.89  & 33.66  & 36.89  & 32.92     \\
    w/o BF.     & 38.09 & 32.31  & 36.58  & 38.40  & 33.13  & 36.76  & 32.81     \\
    w/o CA.     & 35.29 & 29.54  & 31.29  & 35.49  & 31.91  & 35.72  & 31.08      \\
    \midrule
    All        & \textbf{38.91} & \textbf{33.69}  & \textbf{37.56}  & \textbf{39.51}  & \textbf{34.73}  & \textbf{38.66}  & \textbf{34.01}     \\
    \bottomrule
  \end{tabular}
  }
\end{table}

\subsubsection{Parameter Sensitivity (RQ3)}
We further analyze the sensitivity of hyperparameter settings in ~\M. As shown in Figure \ref{fig:parameter}, we select the periods with the lowest (P15) and highest (P17) metrics for display to ensure that our samples are representative. We have chosen the following three hyperparameters, whose detailed descriptions are provided below: (i) $\gamma \Rightarrow$ number of skip steps to accelerate diffusion sampling; (ii) $\xi \Rightarrow$ balancing the weight of the reconstruction of the expert trajectory and action generation within the loss functions; (iii) $\delta \Rightarrow$ harmonizing the expert representation within the blend-forcing paradigm. The experimental results indicate that the parameters $\gamma$ and $\xi$ are significantly influenced by hyperparameter changes, whereas $\delta$ is relatively less affected. Furthermore, it has been demonstrated that employing an appropriate skipping strategy can yield optimal results. Since a larger value of $\xi$ determines the model’s ability to reconstruct the expert and the action, placing greater emphasis on reconstruction may restrict the generalization capabilities of the DDPM, resulting in a gradual decline in performance metrics.

\begin{figure}[ht]
  \centering
  \includegraphics[width=\linewidth]{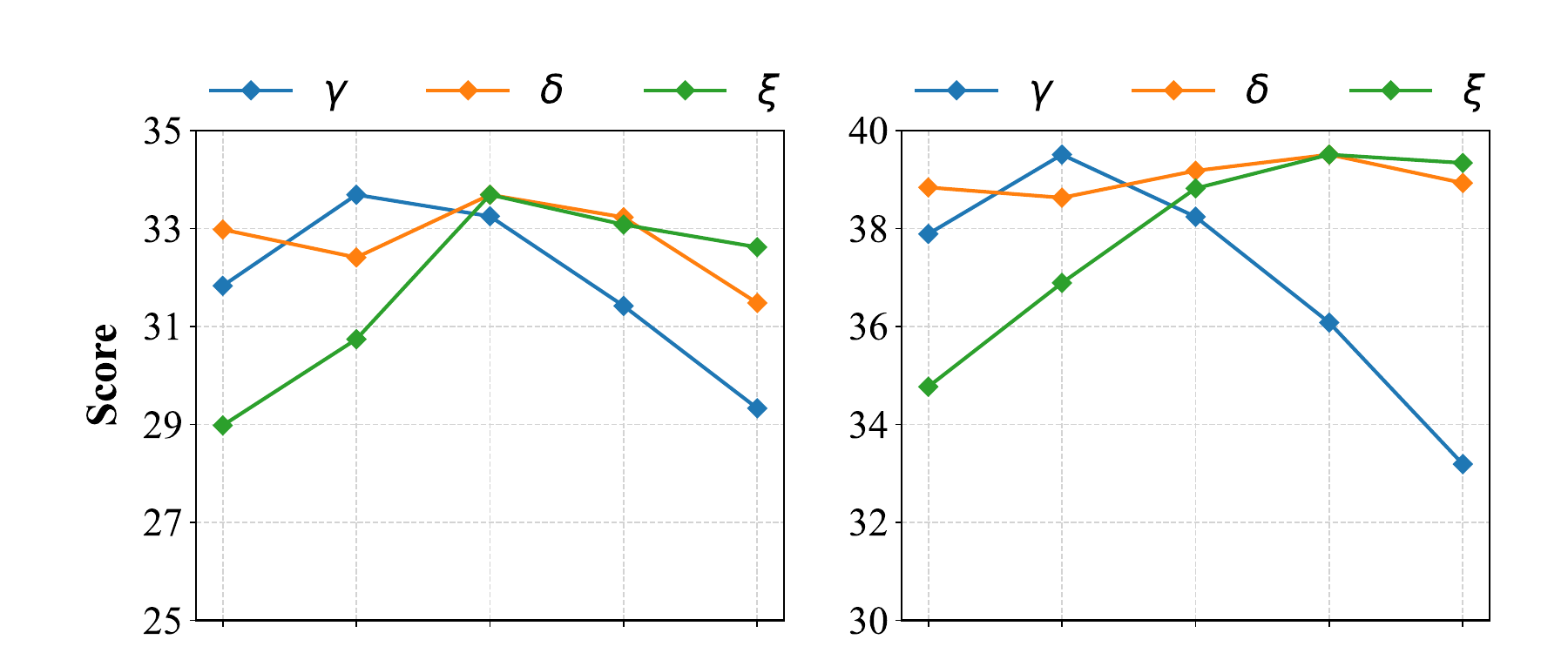}
  \caption{We examine the sensitivity of three key hyperparameters in the period of P15 (left) and P17 (right). The horizontal axis displays $\gamma$ for $\{1,2,4,8,16\}$, $\delta$ for $\{0, 0.2, 0.4, 0.6, 0.8\}$, and $\xi$ for $\{0.25, 0.50, 1.00, 2.00 ,4.00\}$.}
  \label{fig:parameter}
\end{figure}

\subsection{Online A/B Testing \textbf{(RQ4)}}

\begin{figure}[t]
    \centering
    \includegraphics[width=\linewidth]{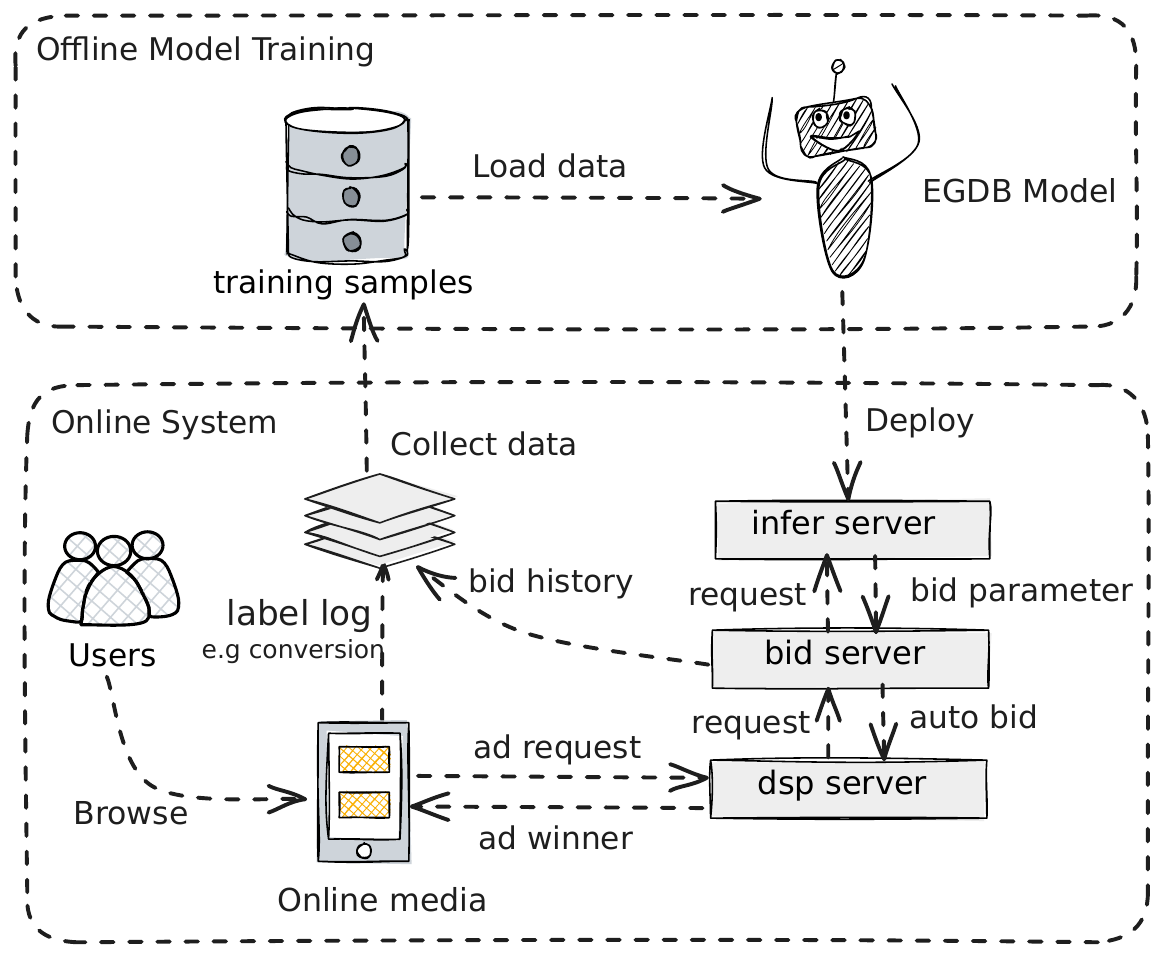}
    \caption{The workflow of proposed ~\M~ in online system.}
    \label{fig:parameter}
\end{figure}

To verify the performance of the proposed method in real-world business scenarios, we selected a batch of the lowest-cost campaigns, aimed at maximizing conversions within budget constraints. DiffBid~\cite{guo2024aigb} serves as a baseline to evaluate the effectiveness of our proposed method, as our modifications are built upon it. We conducted a budget and traffic-splitting online experiment from 7 February 2025 to 13 February 2025, respectively, and the online features and actions can be detailed in the following:

\begin{itemize}[leftmargin=*]
    \item \textbf{Features:} a series of window features with different window sizes, including the bid coefficient, remaining traffic, remaining budget, cumulative consumption, cumulative revenue, etc.
    \item \textbf{Actions:} the difference between two consecutive bid coefficients is represented as $ a_t = b_{t+1} - b_t $, where \( b_t \) denotes the bid coefficient for a specific advertiser at time $t$.
\end{itemize}
The experiment results are presented in Table~\ref{t:online}. Our method demonstrated a significant improvement compared to the baseline model. Specifically, ~\M~ increased conversions by 11.29\% and revenue by 12.35\%. In contrast to DiffBid, \M~ primarily integrates VAE expert trajectory generation and cross-attention-based denoising steps, which influence inference complexity. However, we utilized the skip sampling method to reduce 25\% of the inference resources when serving the same advertising campaign while significantly maintaining the model’s effectiveness.

\begin{table}[ht]
\caption{Performance of ~\M~ in Online A/B Testing.}
\label{t:online}
\resizebox{\linewidth}{!}{
\centering 
\begin{tabular}{lcccc}
\toprule
\textbf{Method} & \textbf{Budget}& \textbf{Campaign} & \textbf{Conversions} & \textbf{Revenue} \\ 
\midrule
DiffBid     & 269,686 &247   & 265,155         & 1,188,336  \\ 
\M~            & 269,686 &247   & 295,090         & 1,335,214  \\ 
\midrule
\textit{Improv.}     & ---     &---     & \tcr{+11.29\%}        & \tcr{+12.36\%}  \\ 
\bottomrule
\end{tabular}
}
\end{table}

\section{Conclusion}
In this paper, we propose \M, an expert-guided conditional diffusion model for auto-bidding. To enhance the diffusion planner with a more robust and personalized condition, we extract expert trajectories and introduce the EGCD block, which integrates expert conditions into diffusion transformers. The inherent behavior of the expert is modeled using the VAE block, and we propose a blended-forcing technique to improve training efficiency. Additionally, we adopt accelerated sampling with a skip-step approach to speed up the generation process for matching online inference. Offline and online experiments have been designed to demonstrate superiority over baseline models. Future work will explore optimal path generation principles and hybrid architectures that combine diffusion models with the advantages of linear programming.

\balance
\bibliographystyle{ACM-Reference-Format}
\bibliography{reference}


\begin{thebibliography}{43}


\ifx \showCODEN    \undefined \def \showCODEN     #1{\unskip}     \fi
\ifx \showDOI      \undefined \def \showDOI       #1{#1}\fi
\ifx \showISBNx    \undefined \def \showISBNx     #1{\unskip}     \fi
\ifx \showISBNxiii \undefined \def \showISBNxiii  #1{\unskip}     \fi
\ifx \showISSN     \undefined \def \showISSN      #1{\unskip}     \fi
\ifx \showLCCN     \undefined \def \showLCCN      #1{\unskip}     \fi
\ifx \shownote     \undefined \def \shownote      #1{#1}          \fi
\ifx \showarticletitle \undefined \def \showarticletitle #1{#1}   \fi
\ifx \showURL      \undefined \def \showURL       {\relax}        \fi
\providecommand\bibfield[2]{#2}
\providecommand\bibinfo[2]{#2}
\providecommand\natexlab[1]{#1}
\providecommand\showeprint[2][]{arXiv:#2}

\bibitem[Adikari and Dutta(2015)]%
        {adikari2015real}
\bibfield{author}{\bibinfo{person}{Shalinda Adikari} {and} \bibinfo{person}{Kaushik Dutta}.} \bibinfo{year}{2015}\natexlab{}.
\newblock \showarticletitle{Real time bidding in online digital advertisement}. In \bibinfo{booktitle}{\emph{New Horizons in Design Science: Broadening the Research Agenda: 10th International Conference, DESRIST 2015, Dublin, Ireland, May 20-22, 2015, Proceedings 10}}. Springer, \bibinfo{pages}{19--38}.
\newblock


\bibitem[Aggarwal et~al\mbox{.}(2024)]%
        {aggarwal2024auto}
\bibfield{author}{\bibinfo{person}{Gagan Aggarwal}, \bibinfo{person}{Ashwinkumar Badanidiyuru}, \bibinfo{person}{Santiago~R Balseiro}, \bibinfo{person}{Kshipra Bhawalkar}, \bibinfo{person}{Yuan Deng}, \bibinfo{person}{Zhe Feng}, \bibinfo{person}{Gagan Goel}, \bibinfo{person}{Christopher Liaw}, \bibinfo{person}{Haihao Lu}, \bibinfo{person}{Mohammad Mahdian}, {et~al\mbox{.}}} \bibinfo{year}{2024}\natexlab{}.
\newblock \showarticletitle{Auto-bidding and auctions in online advertising: A survey}.
\newblock \bibinfo{journal}{\emph{ACM SIGecom Exchanges}} \bibinfo{volume}{22}, \bibinfo{number}{1} (\bibinfo{year}{2024}), \bibinfo{pages}{159--183}.
\newblock


\bibitem[Agrawal et~al\mbox{.}(2016)]%
        {agrawal2016learning}
\bibfield{author}{\bibinfo{person}{Pulkit Agrawal}, \bibinfo{person}{Ashvin~V Nair}, \bibinfo{person}{Pieter Abbeel}, \bibinfo{person}{Jitendra Malik}, {and} \bibinfo{person}{Sergey Levine}.} \bibinfo{year}{2016}\natexlab{}.
\newblock \showarticletitle{Learning to poke by poking: Experiential learning of intuitive physics}.
\newblock \bibinfo{journal}{\emph{Advances in neural information processing systems}}  \bibinfo{volume}{29} (\bibinfo{year}{2016}).
\newblock


\bibitem[Ajay et~al\mbox{.}(2022)]%
        {ajay2022conditional}
\bibfield{author}{\bibinfo{person}{Anurag Ajay}, \bibinfo{person}{Yilun Du}, \bibinfo{person}{Abhi Gupta}, \bibinfo{person}{Joshua Tenenbaum}, \bibinfo{person}{Tommi Jaakkola}, {and} \bibinfo{person}{Pulkit Agrawal}.} \bibinfo{year}{2022}\natexlab{}.
\newblock \showarticletitle{Is conditional generative modeling all you need for decision-making?}
\newblock \bibinfo{journal}{\emph{arXiv preprint arXiv:2211.15657}} (\bibinfo{year}{2022}).
\newblock


\bibitem[Bengio et~al\mbox{.}(2015)]%
        {bengio2015scheduled}
\bibfield{author}{\bibinfo{person}{Samy Bengio}, \bibinfo{person}{Oriol Vinyals}, \bibinfo{person}{Navdeep Jaitly}, {and} \bibinfo{person}{Noam Shazeer}.} \bibinfo{year}{2015}\natexlab{}.
\newblock \showarticletitle{Scheduled sampling for sequence prediction with recurrent neural networks}.
\newblock \bibinfo{journal}{\emph{Advances in neural information processing systems}}  \bibinfo{volume}{28} (\bibinfo{year}{2015}).
\newblock


\bibitem[Cai et~al\mbox{.}(2017)]%
        {cai2017real}
\bibfield{author}{\bibinfo{person}{Han Cai}, \bibinfo{person}{Kan Ren}, \bibinfo{person}{Weinan Zhang}, \bibinfo{person}{Kleanthis Malialis}, \bibinfo{person}{Jun Wang}, \bibinfo{person}{Yong Yu}, {and} \bibinfo{person}{Defeng Guo}.} \bibinfo{year}{2017}\natexlab{}.
\newblock \showarticletitle{Real-time bidding by reinforcement learning in display advertising}. In \bibinfo{booktitle}{\emph{Proceedings of the tenth ACM international conference on web search and data mining}}. \bibinfo{pages}{661--670}.
\newblock


\bibitem[Chen et~al\mbox{.}(2022)]%
        {chen2022offline}
\bibfield{author}{\bibinfo{person}{Huayu Chen}, \bibinfo{person}{Cheng Lu}, \bibinfo{person}{Chengyang Ying}, \bibinfo{person}{Hang Su}, {and} \bibinfo{person}{Jun Zhu}.} \bibinfo{year}{2022}\natexlab{}.
\newblock \showarticletitle{Offline reinforcement learning via high-fidelity generative behavior modeling}.
\newblock \bibinfo{journal}{\emph{arXiv preprint arXiv:2209.14548}} (\bibinfo{year}{2022}).
\newblock


\bibitem[Chen et~al\mbox{.}(2021)]%
        {dt}
\bibfield{author}{\bibinfo{person}{Lili Chen}, \bibinfo{person}{Kevin Lu}, \bibinfo{person}{Aravind Rajeswaran}, \bibinfo{person}{Kimin Lee}, \bibinfo{person}{Aditya Grover}, \bibinfo{person}{Michael Laskin}, \bibinfo{person}{Pieter Abbeel}, \bibinfo{person}{Aravind Srinivas}, {and} \bibinfo{person}{Igor Mordatch}.} \bibinfo{year}{2021}\natexlab{}.
\newblock \showarticletitle{Decision Transformer: Reinforcement Learning via Sequence Modeling}. In \bibinfo{booktitle}{\emph{NeurIPS}}.
\newblock


\bibitem[Evans(2009)]%
        {evans2009online}
\bibfield{author}{\bibinfo{person}{David~S Evans}.} \bibinfo{year}{2009}\natexlab{}.
\newblock \showarticletitle{The online advertising industry: Economics, evolution, and privacy}.
\newblock \bibinfo{journal}{\emph{Journal of economic perspectives}} \bibinfo{volume}{23}, \bibinfo{number}{3} (\bibinfo{year}{2009}), \bibinfo{pages}{37--60}.
\newblock


\bibitem[Fu et~al\mbox{.}(2024)]%
        {fu2024unveil}
\bibfield{author}{\bibinfo{person}{Hengyu Fu}, \bibinfo{person}{Zhuoran Yang}, \bibinfo{person}{Mengdi Wang}, {and} \bibinfo{person}{Minshuo Chen}.} \bibinfo{year}{2024}\natexlab{}.
\newblock \showarticletitle{Unveil conditional diffusion models with classifier-free guidance: A sharp statistical theory}.
\newblock \bibinfo{journal}{\emph{arXiv preprint arXiv:2403.11968}} (\bibinfo{year}{2024}).
\newblock


\bibitem[Fujimoto et~al\mbox{.}(2019)]%
        {bcq}
\bibfield{author}{\bibinfo{person}{Scott Fujimoto}, \bibinfo{person}{David Meger}, {and} \bibinfo{person}{Doina Precup}.} \bibinfo{year}{2019}\natexlab{}.
\newblock \showarticletitle{Off-policy deep reinforcement learning without exploration}. In \bibinfo{booktitle}{\emph{International conference on machine learning}}. PMLR, \bibinfo{pages}{2052--2062}.
\newblock


\bibitem[Gao et~al\mbox{.}(2025)]%
        {gao2025generative}
\bibfield{author}{\bibinfo{person}{Jingtong Gao}, \bibinfo{person}{Yewen Li}, \bibinfo{person}{Shuai Mao}, \bibinfo{person}{Peng Jiang}, \bibinfo{person}{Nan Jiang}, \bibinfo{person}{Yejing Wang}, \bibinfo{person}{Qingpeng Cai}, \bibinfo{person}{Fei Pan}, \bibinfo{person}{Kun Gai}, \bibinfo{person}{Bo An}, {et~al\mbox{.}}} \bibinfo{year}{2025}\natexlab{}.
\newblock \showarticletitle{Generative Auto-Bidding with Value-Guided Explorations}.
\newblock \bibinfo{journal}{\emph{arXiv preprint arXiv:2504.14587}} (\bibinfo{year}{2025}).
\newblock


\bibitem[Guo et~al\mbox{.}(2024a)]%
        {guo2024aigb}
\bibfield{author}{\bibinfo{person}{Jiayan Guo}, \bibinfo{person}{Yusen Huo}, \bibinfo{person}{Zhilin Zhang}, \bibinfo{person}{Tianyu Wang}, \bibinfo{person}{Chuan Yu}, \bibinfo{person}{Jian Xu}, \bibinfo{person}{Yan Zhang}, {and} \bibinfo{person}{Bo Zheng}.} \bibinfo{year}{2024}\natexlab{a}.
\newblock \showarticletitle{AIGB: Generative Auto-bidding via Diffusion Modeling}.
\newblock \bibinfo{journal}{\emph{arXiv preprint arXiv:2405.16141}} (\bibinfo{year}{2024}).
\newblock


\bibitem[Guo et~al\mbox{.}(2024b)]%
        {guo2024generative}
\bibfield{author}{\bibinfo{person}{Jiayan Guo}, \bibinfo{person}{Yusen Huo}, \bibinfo{person}{Zhilin Zhang}, \bibinfo{person}{Tianyu Wang}, \bibinfo{person}{Chuan Yu}, \bibinfo{person}{Jian Xu}, \bibinfo{person}{Bo Zheng}, {and} \bibinfo{person}{Yan Zhang}.} \bibinfo{year}{2024}\natexlab{b}.
\newblock \showarticletitle{Generative Auto-bidding via Conditional Diffusion Modeling}. In \bibinfo{booktitle}{\emph{Proceedings of the 30th ACM SIGKDD Conference on Knowledge Discovery and Data Mining}}. \bibinfo{pages}{5038--5049}.
\newblock


\bibitem[Hansen-Estruch et~al\mbox{.}(2023)]%
        {hansen2023idql}
\bibfield{author}{\bibinfo{person}{Philippe Hansen-Estruch}, \bibinfo{person}{Ilya Kostrikov}, \bibinfo{person}{Michael Janner}, \bibinfo{person}{Jakub~Grudzien Kuba}, {and} \bibinfo{person}{Sergey Levine}.} \bibinfo{year}{2023}\natexlab{}.
\newblock \showarticletitle{Idql: Implicit q-learning as an actor-critic method with diffusion policies}.
\newblock \bibinfo{journal}{\emph{arXiv preprint arXiv:2304.10573}} (\bibinfo{year}{2023}).
\newblock


\bibitem[He et~al\mbox{.}(2021)]%
        {he2021unified}
\bibfield{author}{\bibinfo{person}{Yue He}, \bibinfo{person}{Xiujun Chen}, \bibinfo{person}{Di Wu}, \bibinfo{person}{Junwei Pan}, \bibinfo{person}{Qing Tan}, \bibinfo{person}{Chuan Yu}, \bibinfo{person}{Jian Xu}, {and} \bibinfo{person}{Xiaoqiang Zhu}.} \bibinfo{year}{2021}\natexlab{}.
\newblock \showarticletitle{A unified solution to constrained bidding in online display advertising}. In \bibinfo{booktitle}{\emph{Proceedings of the 27th ACM SIGKDD Conference on Knowledge Discovery \& Data Mining}}. \bibinfo{pages}{2993--3001}.
\newblock


\bibitem[Heng et~al\mbox{.}(2024)]%
        {heng2024out}
\bibfield{author}{\bibinfo{person}{Alvin Heng}, \bibinfo{person}{Harold Soh}, {et~al\mbox{.}}} \bibinfo{year}{2024}\natexlab{}.
\newblock \showarticletitle{Out-of-distribution detection with a single unconditional diffusion model}.
\newblock \bibinfo{journal}{\emph{Advances in Neural Information Processing Systems}}  \bibinfo{volume}{37} (\bibinfo{year}{2024}), \bibinfo{pages}{43952--43974}.
\newblock


\bibitem[Ho et~al\mbox{.}(2020)]%
        {ho2020denoising}
\bibfield{author}{\bibinfo{person}{Jonathan Ho}, \bibinfo{person}{Ajay Jain}, {and} \bibinfo{person}{Pieter Abbeel}.} \bibinfo{year}{2020}\natexlab{}.
\newblock \showarticletitle{Denoising diffusion probabilistic models}.
\newblock \bibinfo{journal}{\emph{Advances in neural information processing systems}}  \bibinfo{volume}{33} (\bibinfo{year}{2020}), \bibinfo{pages}{6840--6851}.
\newblock


\bibitem[Ho and Salimans(2022)]%
        {ho2022classifier}
\bibfield{author}{\bibinfo{person}{Jonathan Ho} {and} \bibinfo{person}{Tim Salimans}.} \bibinfo{year}{2022}\natexlab{}.
\newblock \showarticletitle{Classifier-free diffusion guidance}.
\newblock \bibinfo{journal}{\emph{arXiv preprint arXiv:2207.12598}} (\bibinfo{year}{2022}).
\newblock


\bibitem[Huang et~al\mbox{.}(2024)]%
        {huang2024dual}
\bibfield{author}{\bibinfo{person}{Hongtao Huang}, \bibinfo{person}{Chengkai Huang}, \bibinfo{person}{Xiaojun Chang}, \bibinfo{person}{Wen Hu}, {and} \bibinfo{person}{Lina Yao}.} \bibinfo{year}{2024}\natexlab{}.
\newblock \showarticletitle{Dual Conditional Diffusion Models for Sequential Recommendation}.
\newblock \bibinfo{journal}{\emph{arXiv preprint arXiv:2410.21967}} (\bibinfo{year}{2024}).
\newblock


\bibitem[Jin et~al\mbox{.}(2018)]%
        {jin2018real}
\bibfield{author}{\bibinfo{person}{Junqi Jin}, \bibinfo{person}{Chengru Song}, \bibinfo{person}{Han Li}, \bibinfo{person}{Kun Gai}, \bibinfo{person}{Jun Wang}, {and} \bibinfo{person}{Weinan Zhang}.} \bibinfo{year}{2018}\natexlab{}.
\newblock \showarticletitle{Real-time bidding with multi-agent reinforcement learning in display advertising}. In \bibinfo{booktitle}{\emph{Proceedings of the 27th ACM international conference on information and knowledge management}}. \bibinfo{pages}{2193--2201}.
\newblock


\bibitem[Johnson et~al\mbox{.}(2020)]%
        {johnson2020consumer}
\bibfield{author}{\bibinfo{person}{Garrett~A Johnson}, \bibinfo{person}{Scott~K Shriver}, {and} \bibinfo{person}{Shaoyin Du}.} \bibinfo{year}{2020}\natexlab{}.
\newblock \showarticletitle{Consumer privacy choice in online advertising: Who opts out and at what cost to industry?}
\newblock \bibinfo{journal}{\emph{Marketing Science}} \bibinfo{volume}{39}, \bibinfo{number}{1} (\bibinfo{year}{2020}), \bibinfo{pages}{33--51}.
\newblock


\bibitem[Kingma(2013)]%
        {kingma2013auto}
\bibfield{author}{\bibinfo{person}{Diederik~P Kingma}.} \bibinfo{year}{2013}\natexlab{}.
\newblock \showarticletitle{Auto-encoding variational bayes}.
\newblock \bibinfo{journal}{\emph{arXiv preprint arXiv:1312.6114}} (\bibinfo{year}{2013}).
\newblock


\bibitem[Kostrikov et~al\mbox{.}(2022)]%
        {iql}
\bibfield{author}{\bibinfo{person}{Ilya Kostrikov}, \bibinfo{person}{Ashvin Nair}, {and} \bibinfo{person}{Sergey Levine}.} \bibinfo{year}{2022}\natexlab{}.
\newblock \showarticletitle{Offline Reinforcement Learning with Implicit Q-Learning}. In \bibinfo{booktitle}{\emph{ICLR}}.
\newblock


\bibitem[Kumar et~al\mbox{.}(2020)]%
        {cql}
\bibfield{author}{\bibinfo{person}{Aviral Kumar}, \bibinfo{person}{Aurick Zhou}, \bibinfo{person}{George Tucker}, {and} \bibinfo{person}{Sergey Levine}.} \bibinfo{year}{2020}\natexlab{}.
\newblock \showarticletitle{Conservative Q-Learning for Offline Reinforcement Learning}. In \bibinfo{booktitle}{\emph{NeurIPS}}.
\newblock


\bibitem[Li et~al\mbox{.}(2025)]%
        {li2025ebaret}
\bibfield{author}{\bibinfo{person}{Kaiyuan Li}, \bibinfo{person}{Pengyu Wang}, \bibinfo{person}{Yunshan Peng}, \bibinfo{person}{Pengjia Yuan}, \bibinfo{person}{Yanxiang Zeng}, \bibinfo{person}{Rui Xiang}, \bibinfo{person}{Yanhua Cheng}, \bibinfo{person}{Xialong Liu}, {and} \bibinfo{person}{Peng Jiang}.} \bibinfo{year}{2025}\natexlab{}.
\newblock \showarticletitle{EBaReT: Expert-guided Bag Reward Transformer for Auto Bidding}. In \bibinfo{booktitle}{\emph{Companion Proceedings of the ACM on Web Conference 2025}}. \bibinfo{pages}{1104--1108}.
\newblock


\bibitem[Li et~al\mbox{.}(2024)]%
        {li2024gas}
\bibfield{author}{\bibinfo{person}{Yewen Li}, \bibinfo{person}{Shuai Mao}, \bibinfo{person}{Jingtong Gao}, \bibinfo{person}{Nan Jiang}, \bibinfo{person}{Yunjian Xu}, \bibinfo{person}{Qingpeng Cai}, \bibinfo{person}{Fei Pan}, \bibinfo{person}{Peng Jiang}, {and} \bibinfo{person}{Bo An}.} \bibinfo{year}{2024}\natexlab{}.
\newblock \showarticletitle{GAS: Generative Auto-bidding with Post-training Search}.
\newblock \bibinfo{journal}{\emph{arXiv preprint arXiv:2412.17018}} (\bibinfo{year}{2024}).
\newblock


\bibitem[Li et~al\mbox{.}(2023)]%
        {li2023diffurec}
\bibfield{author}{\bibinfo{person}{Zihao Li}, \bibinfo{person}{Aixin Sun}, {and} \bibinfo{person}{Chenliang Li}.} \bibinfo{year}{2023}\natexlab{}.
\newblock \showarticletitle{Diffurec: A diffusion model for sequential recommendation}.
\newblock \bibinfo{journal}{\emph{ACM Transactions on Information Systems}} \bibinfo{volume}{42}, \bibinfo{number}{3} (\bibinfo{year}{2023}), \bibinfo{pages}{1--28}.
\newblock


\bibitem[Lin et~al\mbox{.}(2023)]%
        {lin2023safe}
\bibfield{author}{\bibinfo{person}{Qian Lin}, \bibinfo{person}{Bo Tang}, \bibinfo{person}{Zifan Wu}, \bibinfo{person}{Chao Yu}, \bibinfo{person}{Shangqin Mao}, \bibinfo{person}{Qianlong Xie}, \bibinfo{person}{Xingxing Wang}, {and} \bibinfo{person}{Dong Wang}.} \bibinfo{year}{2023}\natexlab{}.
\newblock \showarticletitle{Safe offline reinforcement learning with real-time budget constraints}. In \bibinfo{booktitle}{\emph{International Conference on Machine Learning}}. PMLR, \bibinfo{pages}{21127--21152}.
\newblock


\bibitem[Lin et~al\mbox{.}(2024)]%
        {lin2024robust}
\bibfield{author}{\bibinfo{person}{Qilong Lin}, \bibinfo{person}{Zhenzhe Zheng}, {and} \bibinfo{person}{Fan Wu}.} \bibinfo{year}{2024}\natexlab{}.
\newblock \showarticletitle{Robust Auto-Bidding Strategies for Online Advertising}. In \bibinfo{booktitle}{\emph{Proceedings of the 30th ACM SIGKDD Conference on Knowledge Discovery and Data Mining}}. \bibinfo{pages}{1804--1815}.
\newblock


\bibitem[Mou et~al\mbox{.}(2022)]%
        {mou2022sustainable}
\bibfield{author}{\bibinfo{person}{Zhiyu Mou}, \bibinfo{person}{Yusen Huo}, \bibinfo{person}{Rongquan Bai}, \bibinfo{person}{Mingzhou Xie}, \bibinfo{person}{Chuan Yu}, \bibinfo{person}{Jian Xu}, {and} \bibinfo{person}{Bo Zheng}.} \bibinfo{year}{2022}\natexlab{}.
\newblock \showarticletitle{Sustainable online reinforcement learning for auto-bidding}.
\newblock \bibinfo{journal}{\emph{Advances in Neural Information Processing Systems}}  \bibinfo{volume}{35} (\bibinfo{year}{2022}), \bibinfo{pages}{2651--2663}.
\newblock


\bibitem[Song et~al\mbox{.}(2020)]%
        {song2020denoising}
\bibfield{author}{\bibinfo{person}{Jiaming Song}, \bibinfo{person}{Chenlin Meng}, {and} \bibinfo{person}{Stefano Ermon}.} \bibinfo{year}{2020}\natexlab{}.
\newblock \showarticletitle{Denoising diffusion implicit models}.
\newblock \bibinfo{journal}{\emph{arXiv preprint arXiv:2010.02502}} (\bibinfo{year}{2020}).
\newblock


\bibitem[Tianchi(2024)]%
        {AIGB_data}
\bibfield{author}{\bibinfo{person}{Tianchi}.} \bibinfo{year}{2024}\natexlab{}.
\newblock \bibinfo{booktitle}{\emph{AIGB Track: Learning Auto-Bidding Agent with Generative Models}}.
\newblock
\urldef\tempurl%
\url{https://tianchi.aliyun.com/competition/entrance/532236/customize448?lang=en-us}
\showURL{%
\tempurl}


\bibitem[Torabi et~al\mbox{.}(2018)]%
        {bc}
\bibfield{author}{\bibinfo{person}{Faraz Torabi}, \bibinfo{person}{Garrett Warnell}, {and} \bibinfo{person}{Peter Stone}.} \bibinfo{year}{2018}\natexlab{}.
\newblock \showarticletitle{Behavioral Cloning from Observation}. In \bibinfo{booktitle}{\emph{IJCAI}}, \bibfield{editor}{\bibinfo{person}{J{\'{e}}r{\^{o}}me Lang}} (Ed.). \bibinfo{pages}{4950--4957}.
\newblock


\bibitem[Vaswani et~al\mbox{.}(2017)]%
        {vaswani2017attention}
\bibfield{author}{\bibinfo{person}{Ashish Vaswani}, \bibinfo{person}{Noam Shazeer}, \bibinfo{person}{Niki Parmar}, \bibinfo{person}{Jakob Uszkoreit}, \bibinfo{person}{Llion Jones}, \bibinfo{person}{Aidan~N Gomez}, \bibinfo{person}{{\L}ukasz Kaiser}, {and} \bibinfo{person}{Illia Polosukhin}.} \bibinfo{year}{2017}\natexlab{}.
\newblock \showarticletitle{Attention is all you need}.
\newblock \bibinfo{journal}{\emph{Advances in neural information processing systems}}  \bibinfo{volume}{30} (\bibinfo{year}{2017}).
\newblock


\bibitem[Wang et~al\mbox{.}(2017)]%
        {wang2017display}
\bibfield{author}{\bibinfo{person}{Jun Wang}, \bibinfo{person}{Weinan Zhang}, \bibinfo{person}{Shuai Yuan}, {et~al\mbox{.}}} \bibinfo{year}{2017}\natexlab{}.
\newblock \showarticletitle{Display advertising with real-time bidding (RTB) and behavioural targeting}.
\newblock \bibinfo{journal}{\emph{Foundations and Trends{\textregistered} in Information Retrieval}} \bibinfo{volume}{11}, \bibinfo{number}{4-5} (\bibinfo{year}{2017}), \bibinfo{pages}{297--435}.
\newblock


\bibitem[Wang et~al\mbox{.}(2024)]%
        {wang2024conditional}
\bibfield{author}{\bibinfo{person}{Yu Wang}, \bibinfo{person}{Zhiwei Liu}, \bibinfo{person}{Liangwei Yang}, {and} \bibinfo{person}{Philip~S Yu}.} \bibinfo{year}{2024}\natexlab{}.
\newblock \showarticletitle{Conditional denoising diffusion for sequential recommendation}. In \bibinfo{booktitle}{\emph{Pacific-Asia Conference on Knowledge Discovery and Data Mining}}. Springer, \bibinfo{pages}{156--169}.
\newblock


\bibitem[Wang et~al\mbox{.}(2022)]%
        {wang2022diffusion}
\bibfield{author}{\bibinfo{person}{Zhendong Wang}, \bibinfo{person}{Jonathan~J Hunt}, {and} \bibinfo{person}{Mingyuan Zhou}.} \bibinfo{year}{2022}\natexlab{}.
\newblock \showarticletitle{Diffusion policies as an expressive policy class for offline reinforcement learning}.
\newblock \bibinfo{journal}{\emph{arXiv preprint arXiv:2208.06193}} (\bibinfo{year}{2022}).
\newblock


\bibitem[Wen et~al\mbox{.}(2022)]%
        {wen2022cooperative}
\bibfield{author}{\bibinfo{person}{Chao Wen}, \bibinfo{person}{Miao Xu}, \bibinfo{person}{Zhilin Zhang}, \bibinfo{person}{Zhenzhe Zheng}, \bibinfo{person}{Yuhui Wang}, \bibinfo{person}{Xiangyu Liu}, \bibinfo{person}{Yu Rong}, \bibinfo{person}{Dong Xie}, \bibinfo{person}{Xiaoyang Tan}, \bibinfo{person}{Chuan Yu}, {et~al\mbox{.}}} \bibinfo{year}{2022}\natexlab{}.
\newblock \showarticletitle{A cooperative-competitive multi-agent framework for auto-bidding in online advertising}. In \bibinfo{booktitle}{\emph{Proceedings of the Fifteenth ACM International Conference on Web Search and Data Mining}}. \bibinfo{pages}{1129--1139}.
\newblock


\bibitem[Williams and Zipser(1989)]%
        {williams1989learning}
\bibfield{author}{\bibinfo{person}{Ronald~J Williams} {and} \bibinfo{person}{David Zipser}.} \bibinfo{year}{1989}\natexlab{}.
\newblock \showarticletitle{A learning algorithm for continually running fully recurrent neural networks}.
\newblock \bibinfo{journal}{\emph{Neural computation}} \bibinfo{volume}{1}, \bibinfo{number}{2} (\bibinfo{year}{1989}), \bibinfo{pages}{270--280}.
\newblock


\bibitem[Yang et~al\mbox{.}(2024)]%
        {yang2024generate}
\bibfield{author}{\bibinfo{person}{Zhengyi Yang}, \bibinfo{person}{Jiancan Wu}, \bibinfo{person}{Zhicai Wang}, \bibinfo{person}{Xiang Wang}, \bibinfo{person}{Yancheng Yuan}, {and} \bibinfo{person}{Xiangnan He}.} \bibinfo{year}{2024}\natexlab{}.
\newblock \showarticletitle{Generate what you prefer: Reshaping sequential recommendation via guided diffusion}.
\newblock \bibinfo{journal}{\emph{Advances in Neural Information Processing Systems}}  \bibinfo{volume}{36} (\bibinfo{year}{2024}).
\newblock


\bibitem[Yuan et~al\mbox{.}(2013)]%
        {yuan2013real}
\bibfield{author}{\bibinfo{person}{Shuai Yuan}, \bibinfo{person}{Jun Wang}, {and} \bibinfo{person}{Xiaoxue Zhao}.} \bibinfo{year}{2013}\natexlab{}.
\newblock \showarticletitle{Real-time bidding for online advertising: measurement and analysis}. In \bibinfo{booktitle}{\emph{Proceedings of the seventh international workshop on data mining for online advertising}}. \bibinfo{pages}{1--8}.
\newblock


\bibitem[Zhu et~al\mbox{.}(2023)]%
        {zhu2023diffusion}
\bibfield{author}{\bibinfo{person}{Zhengbang Zhu}, \bibinfo{person}{Hanye Zhao}, \bibinfo{person}{Haoran He}, \bibinfo{person}{Yichao Zhong}, \bibinfo{person}{Shenyu Zhang}, \bibinfo{person}{Haoquan Guo}, \bibinfo{person}{Tingting Chen}, {and} \bibinfo{person}{Weinan Zhang}.} \bibinfo{year}{2023}\natexlab{}.
\newblock \showarticletitle{Diffusion models for reinforcement learning: A survey}.
\newblock \bibinfo{journal}{\emph{arXiv preprint arXiv:2311.01223}} (\bibinfo{year}{2023}).
\newblock


\end{thebibliography}

\end{document}